\documentclass[10pt,twocolumn,letterpaper]{article}

\usepackage{wacv}
\usepackage{times}
\usepackage{epsfig}
\usepackage{graphicx}
\usepackage{amsmath}
\usepackage{amssymb}
\usepackage{algorithm}
\usepackage[noend]{algorithmic}
\usepackage{booktabs,subcaption,amsfonts,dcolumn}
\usepackage{float}

\newcommand\NoDo{\renewcommand\algorithmicdo{}}

\usepackage[pagebackref=true,breaklinks=true,letterpaper=true,colorlinks,bookmarks=false]{hyperref}

\wacvfinalcopy 

\def\wacvPaperID{319} 

\ifwacvfinal\fi
\begin{document}

\title{3D Semi-Supervised Learning with Uncertainty-Aware Multi-View Co-Training}

\author{Yingda Xia\textsuperscript{1}\thanks{Work done during an internship at Nvidia} , Fengze Liu\textsuperscript{1}, Dong Yang\textsuperscript{2}, Jinzheng Cai\textsuperscript{3}, Lequan Yu\textsuperscript{4}, \\Zhuotun Zhu\textsuperscript{1}, Daguang Xu\textsuperscript{2}, Alan Yuille\textsuperscript{1}, Holger Roth\textsuperscript{2}
\vspace{0.2cm}
\\
\textsuperscript{1}Johns Hopkins University\quad\textsuperscript{2}NVIDIA\\ \quad\textsuperscript{3}University of Florida\quad\textsuperscript{4}The Chinese University of Hong Kong\\}

\maketitle

\begin{abstract}

While making a tremendous impact in various fields, deep neural networks usually require large amounts of labeled data for training which are expensive to collect in many applications, especially in the medical domain.  Unlabeled data, on the other hand, is much more abundant. Semi-supervised learning techniques, such as co-training, could provide a powerful tool to leverage unlabeled data. In this paper, we propose a novel framework, \textbf{uncertainty-aware multi-view co-training} (UMCT), to address semi-supervised learning on 3D data, such as volumetric data from medical imaging. 
In our work, co-training is achieved by exploiting multi-viewpoint consistency of 3D data. We generate different views by rotating or permuting the 3D data and utilize asymmetrical 3D kernels to encourage diversified features in different sub-networks. In addition, we propose an uncertainty-weighted label fusion mechanism to estimate the reliability of each view's prediction with Bayesian deep learning. As one view requires the supervision from other views in co-training, our self-adaptive approach computes a confidence score for the prediction of each unlabeled sample in order to assign a reliable pseudo label. Thus, our approach can take advantage of unlabeled data during training. We show the effectiveness of our proposed semi-supervised method on several public datasets from medical image segmentation tasks (NIH pancreas \& LiTS liver tumor dataset). Meanwhile, a fully-supervised method based on our approach achieved state-of-the-art performances on both the LiTS liver tumor segmentation and the Medical Segmentation Decathlon (MSD) challenge, demonstrating the robustness and value of our framework, even when fully supervised training is feasible.


\end{abstract}

\begin{figure*}[!t]
\begin{center}
    \includegraphics[width=17.5cm]{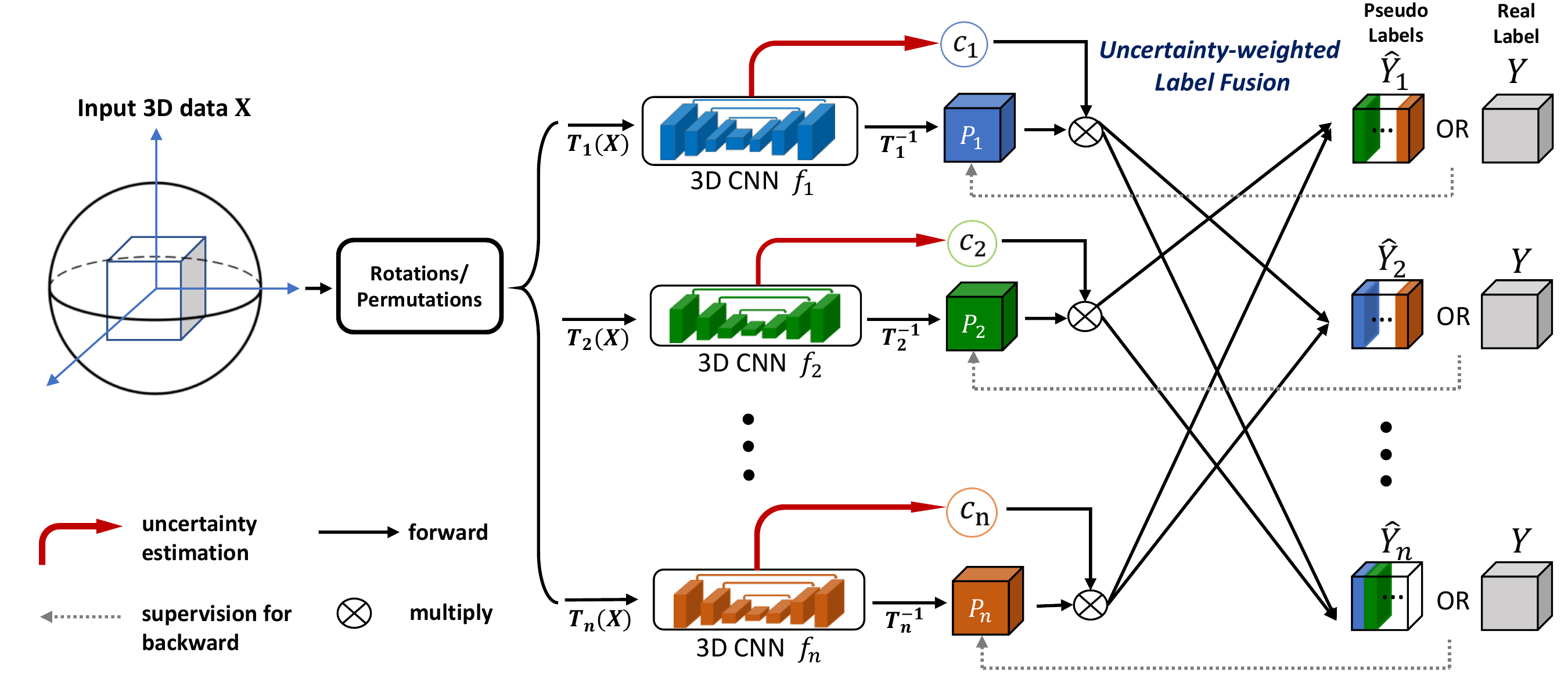}
\end{center}
\vspace{-0.5cm}
\caption{
    Overall framework of \textbf{uncertainty-aware multi-view co-training} (UMCT), best viewed in color. The multi-view inputs of $\mathbf{X}$ are first generated through different transforms $\mathbf{T}$, like rotations and permutations, before being fed into $n$ deep networks with asymmetrical 3D kernels. A confidence score $c$ is computed for each view by uncertainty estimation and acts as the weights to compute the pseudo labels $\hat{Y}$ of other views (Eq. \ref{Eqn:pseudo-label}) after inverse transform $\mathbf{T}^{-1}$ of the predictions. The pseudo labels $\hat{Y}$ for unlabeled data and ground truth $Y$ for labeled data are used as supervisions during training.
}
\label{Fig:Overall}
\vspace{-0.1cm}
\end{figure*}

\section{Introduction}
Deep learning has achieved great successes in various computer vision tasks, such as 2D image recognition ~\cite{krizhevsky2012imagenet,simonyan2014very,szegedy2015going,he2016deep,huang2017densely} and semantic segmentation ~\cite{long2015fully,chen2018deeplab,zhao2017pyramid,chen2017rethinking}. However, deep networks usually rely on large-scale labeled datasets for training. When it comes to 3D data, such as medical volumetric data and point clouds, human labeling can be extremely costly, and often requires expert knowledge. Take medical imaging for example. With the rapid growth in the demand of finer and larger scale of computer-aided diagnoses (CAD), 3D segmentation of medical images (such as CTs and MRIs) is acting as a critical step in biomedical image analysis and surgical planning. However, well-annotated segmentation labels in medical images require both high-level expertise of radiologists and careful manual labeling of object masks or surface boundaries. Therefore, semi-supervised approaches with unlabeled data occupying a large portion of the training data are worth exploring. 

In this paper, we aim to design a semi-supervised approach for 3D data, which can be applied to diverse data sources, e.g. CT/MRI volumes and 3D point clouds. Inspired by the success of co-training ~\cite{blum1998combining} and its extension into single 2D images~\cite{qiao2018deep}, we further extend this idea into 3D. Typical co-training requires at least two views (i.e. sources) of data, either of which should be sufficient to train a classifier on. Co-training minimizes the disagreements by assigning pseudo labels between each other view on unlabeled data. Blum and Mitchell~\cite{blum1998combining} further proved that co-training has PAC-like guarantees on semi-supervised learning with an additional assumption that the two views are conditionally independent given the category. Since most computer vision tasks have only one source of data, encouraging view differences is a crucial point for successful co-training. For example, \textit{deep co-training} ~\cite{qiao2018deep} trains multiple deep networks to act as different views by utilizing adversarial examples ~\cite{Goodfellow2015} to address this issue. Another aspect of co-training to emphasize is view confidence estimation. In multi-view settings, given sufficient variance of each view, the quality of each prediction is not guaranteed and bad pseudo labels can be harmful if used in the training process. Co-training could benefit from trusting reliable predictions and degrading the unreliable ones. However, distinguishing reliable and unreliable predictions is challenging for unlabeled data because of lacking ground-truth.

To address the above two important aspects, we propose an \textit{uncertainty-aware multi-view co-training} (UMCT) framework, shown in Fig.~\ref{Fig:Overall}. First of all, we define the concept of ``view'' in our work as a data-model combination which combines the concepts of data source (classical co-training) and deep network model (deep co-training). Although only one source of data is available, we can still introduce data-level view differences by exploring multiple viewpoints of 3D data through spatial transformations, such as rotation and permutation. Hence, our multi-view approach naturally adapts to analyze 3D data and can be integrated with the proposed co-training framework.

We further introduce the model-level view differences by adopting 2D pre-trained models to asymmetric kernels in 3D networks, such as $3\times3\times1$ kernels. In this way, we can not only utilize the 2D pre-trained weights but also train the whole framework in a full 3D fashion \cite{liu20173d}. Importantly, such design introduces 2D biases in each view during training, leading to complementary feature representations in different views. During the training process, these disagreements between views are minimized through 3D co-training, which further boosts the performance of our model.

Another key component is the view confidence estimation. We propose to estimate the uncertainty of each view's prediction with Bayesian deep networks by adding dropout into the architectures \cite{gal2016dropout}. A confidence score is computed based on epistemic uncertainty \cite{kendall2017uncertainties}, which can act as a weight for each prediction. After propagation through this \textit{uncertainty-weighted label fusion module} (ULF), a set of more accurate pseudo labels can be obtained for each view, which is used as supervision signal for unlabeled data. 

Our proposed approach is evaluated on the NIH pancreas segmentation dataset and the training/validation set of LiTS liver tumor segmentation challenge. It outperforms other semi-supervised methods by a large margin. We further investigate the influence of our approach when applied in a fully supervised setting, to see whether it can also assist training for each branch with sufficient labeled data. A fully-supervised method based on our approach achieved state-of-the-art results on LiTS liver tumor segmentation challenge and scored the second place in the Medical Segmentation Decathlon challenge, without using complicated data augmentation or model ensembles.

\section{Related Work}
\paragraph{Semi-supervised learning.}
Semi-supervised learning approaches aim at learning models with limited labeled data and a large proportion of unlabeled data~\cite{blum1998combining, zhou2005tri, belkin2006manifold, zhou2005semi}. Emerging semi-supervised approaches have been successfully applied to image recognition using deep neural networks ~\cite{laine2016temporal,rasmus2015semi,miyato2018virtual,bachman2014learning,sajjadi2016regularization, chen2018tri}. These algorithms mostly rely on additional regularization terms to train the networks to be resistant to some specific noise. A recent approach ~\cite{qiao2018deep} extended the co-training strategy to 2D deep networks and multiple views, using adversarial examples to encourage view differences to boost performance. 

\vspace{-0.35cm}
\paragraph{Semi-supervised medical image analysis.}~ Cheplygina et al. \cite{cheplygina2018not} mentioned that current semi-supervised medical analysis methods fall into 3 types - self-training (teacher-student models), co-training (with hand-crafted features) and graph-based approaches (mostly applications of graph-cut optimization). Bai et al. ~\cite{bai2017semi} introduced a deep network based self-training framework with conditional random field (CRF) based iterative refinements for medical image segmentation. Zhou et al. ~\cite{zhou2018semi} trained three 2D networks from three planar slices of the 3D data and fused them in each self-training iteration to get a stronger student model. Li et al. ~\cite{li2018semi,li2019transformation} extended the self-ensemble approach $\pi$ model ~\cite{laine2016temporal} with 90-degree rotations making the network rotation-invariant. Generative adversarial network (GAN) based approaches are also popular recently for medical imaging~\cite{mic1,mic2,mic3}.

\vspace{-0.35cm}
\paragraph{Uncertainty estimation.} 
Traditional approaches include particle filtering and CRFs~\cite{blake1993framework,he2004multiscale}. For deep learning, uncertainty is more often measured with Bayesian deep networks~\cite{gal2016dropout,gal2016uncertainty,kendall2017uncertainties}. In our work, we emphasize the importance of uncertainty estimation in semi-supervised learning, since most of the training data here is not annotated. We propose to estimate the confidence of each view in our co-training framework via Bayesian uncertainty estimation.

\vspace{-0.35cm}
\paragraph{2D/3D hybrid networks.}
2D networks and 3D networks both have their advantages and limitations. The former benefit from 2D pre-trained weights and well-studies architectures in natural image processing, while the latter better explore 3D information utilizing 3D convolutional kernels. ~\cite{xia2018bridging,li2017h} either uses 2D probability maps or 2D feature maps for building 3D models. ~\cite{liu20173d} proposed a 3D architecture which can be initialized by 2D pre-trained models. Moreover, ~\cite{roth2016improving,zhou2017fixed} illustrates the effectiveness of multi-view training on 2D slices, even by simply averaging multi-planar results, indicating complementary latent information exists in the biases of 2D networks. 
This inspired us to train 3D multi-view networks with 2D initializations jointly using an additional loss function for multi-view networks which encourages each network to learn from one another.

\section{Uncertainty-aware Multi-view Co-training}
In this section, we introduce our framework of \textit{uncertainty-aware multi-view co-training} (UMCT). There are two important properties for a successful deep network based co-training: view difference and view reliability. In the following sections, we will explain how they are achieved in our 3D framework: a general mathematical formulation of the approach is shown in Sec \ref{sec:overall}; then we demonstrate how to encourage view differences in Sec~\ref{sec:difference}, and how to compute the confidence of each view by uncertainty estimation in Sec \ref{sec:uncertainty}. 

\subsection{Overall Framework}
\label{sec:overall}
We consider the task of semi-supervised learning for 3D data. Let $\mathcal{S}$ and $\mathcal{U}$ be the labeled and unlabeled dataset, respectively. Let $\mathcal{D} = \mathcal{S} \cup \mathcal{U}$ be the whole provided dataset. We denote each labeled data pair as $(\mathbf{X}, \mathbf{Y}) \in \mathcal{S}$ and unlabeled data as $\mathbf{X} \in \mathcal{U}$. The ground truth $\mathbf{Y}$ can either be a ground truth label (classification tasks) or dense prediction map (segmentation tasks).

Suppose for each input $\mathbf{X}$, we can naturally generate $N$ different views of 3D data by applying a transformation $T_i$ (rotation or permutation), which will result in multi-view inputs $T_i(X)$, $i = 1,...,N$. Such operations will introduce a data-level view difference. $N$ models $f_i(\cdot)$, $i = 1,...,N$ are then trained over each view of data respectively. For $(\mathbf{X}, \mathbf{Y}) \in \mathcal{S}$ , a supervised loss function $\mathcal{L}_{sup}$ is optimized to measure the similarity between the prediction of each view $p_i(\mathbf{X}) = T_i^{-1}\circ f_i\circ T_i(\mathbf{X})$ and $\mathbf{Y}$: 
\begin{equation}
\label{Loss1}
\mathcal{L}_{sup}(\mathbf{X}, \mathbf{Y}) = \sum_{i=1}^N \mathcal{L}(p_i(\mathbf{X}), \mathbf{Y}),
\end{equation}where $\mathcal{L}$ is a standard loss function for a supervised learning task (e.g. classification, or segmentation). For 3D segmentation task, $\{p_i(\mathbf{X})\}_{i=1}^N$ are the corresponding voxel-wise prediction score maps after inverse rotation or permutation.

For unlabeled data, we construct a co-training assumption under a semi-supervised setting. 
The co-training strategy assumes the prediction on each view should reach a consensus. So the prediction of each model can act as a pseudo label to supervise other views in order to learn from unlabeled data. 
However, since the prediction of each view is expected to be diverse after boosting the view differences, the quality of each view's prediction needs to be measured before generating trustworthy pseudo labels. This is accomplished by the \textit{uncertainty-weighted label fusion module} (ULF), which is introduced in Sec~\ref{sec:uncertainty}. With ULF, the co-training loss for unlabeled data can be formulated as:
\begin{equation}
\label{Loss2}
\mathcal{L}_{cot}(\mathbf{X}) = \sum_i^N \mathcal{L}(p_i(\mathbf{X}), \hat{\mathbf{Y_i}}), 
\end{equation}
where
\begin{equation}
    \hat{\mathbf{Y_i}} = U_{f_1,..f_n}(p_1(\mathbf{X}),..,p_{i-1}(\mathbf{X}),p_{i+1}(\mathbf{X}),..,p_n(\mathbf{X}))
\label{eqn:pl}
\end{equation}
 is the pseudo label for the $i^{th}$ view, $U_{f_1,..f_n}$ is the ULF computational function.

Overall, the combined loss function is:
\begin{equation}
\label{loss3}
\sum_{(\mathbf{X},\mathbf{Y}) \in \mathcal{S}} \mathcal{L}_{sup}(\mathbf{X}, \mathbf{Y}) + \lambda_{cot} \sum_{\mathbf{X}\in\mathcal{U}}\mathcal{L}_{cot}(\mathbf{X}).
\end{equation}
where $\lambda_{cot}$ is a weight coefficient.

\begin{algorithm}
\caption{Uncertainty-aware Multi-view Co-training}
\textbf{Input:}\\ 
Labeled dataset $\mathcal{S}$ \& Unlabeled dataset $\mathcal{U}$\\
\textit{uncertainty-weighted label fusion module} (ULF) $U_{f_1,..f_n}(\cdot)$\\
\textbf{Output:}\\
Model of each view $f_1,..f_n$
\begin{algorithmic}[1]
\NoDo
\WHILE{stopping criterion not met:}
\STATE Sample batch $b_l = (x_l, y_l) \in \mathcal{S}$ and batch $b_u=(x_u) \in \mathcal{U}$
\STATE Generate multi-view inputs $T_i(x_l)$ and $T_i(x_u)$, $i \in \{1, .., N\}$
\FOR{i \textbf{in} all views:}
\STATE Compute predictions for each view and apply inverse rotation or permutation\\ $p_i(x_l) \gets T_i^{-1}\circ f_i\circ T_i(x_l)$\\$p_i(x_u) \gets T_i^{-1}\circ f_i\circ T_i(x_u)$
\ENDFOR
\FOR{i \textbf{in} all views:}
\STATE Compute pseudo labels for $x_u$ with ULF \\ $\hat{y_i} \gets U_{f_1,..f_n}(p_1(x_u),..,p_{i-1}(x_u),$ \\
         \hspace{3.3cm} $p_{i+1}(x_u),..,p_n(x_u)) $
\ENDFOR
\STATE $\mathcal{L}_{sup} = \frac{1}{|b_l|}\sum_{(x_l,y_l)\in b_l}[\sum_i^N\mathcal{L}(p_i(x_l), y_l)]$
\STATE $\mathcal{L}_{cot} = \frac{1}{|b_u|}\sum_{(x_u)\in b_u}[\sum_i^N\mathcal{L}(p_i(x_u), \hat{y_i})]$
\STATE $\mathcal{L} = \mathcal{L}_{sup} + \lambda_{cot}\mathcal{L}_{cot}$
\STATE Compute gradient of loss function $\mathcal{L}$ and update network parameters $\{\theta_i\}$ by back propagation
\ENDWHILE
\RETURN $f_1,..f_n$
\end{algorithmic}
\label{algo1}
\end{algorithm}

\subsection{Encouraging View Differences}
\label{sec:difference}
A successful co-training requires the ``views" to be different and learn complementary information in the training procedure. 
In our framework, several techniques are proposed to encourage view differences, including both data-level and feature-level.

\vspace{-0.5cm}
\paragraph{3D multi-view generation.} As stated above, in order to generate multi-view data, we transpose $\mathbf{X}$ into multiple views by rotations or permutations $\mathbf{T}$.  (A permutation rearranges the dimensions of an array in a specific order.)  For three-view co-training, these can correspond to the coronal, sagittal and axial views in medical imaging, which matches the multi-planar reformatted views that radiologists typically use to analyze the image. Such operation is a natural way to introduce data-level view difference.

\vspace{-0.5cm}
\paragraph{Asymmetric 3D kernels and 2D initialization.}
The co-training assumption encourages models to make similar predictions on both $\mathcal{S}$ and $\mathcal{U}$, which potentially can lead to collapsed neural networks mentioned in ~\cite{qiao2018deep}. To address this problem, we further encourage view difference at feature level by designing a task-specific model. We propose to use asymmetric 3D models initialized with 2D pre-trained weights as the backbone network of each view to encourage diverse features for each view learning. The simplest version of an asymmetric 3D model is to use $n\times n\times 1$ convolutional kernels instead of $n\times n \times n$ 3D kernels as in common 3D networks. This structure also makes the model convenient to be initialized with 2D pre-trained weights but fine-tuned in a 3D fashion  \cite{liu20173d}. 

\subsection{Compute Reliable Psuedo Labels for Unlabeled Data with Uncertainty Estimation}
\label{sec:uncertainty}
Encouraging view difference means enlarging the variance of each view's prediction $var(p_i(\mathbf{X}))$. This raises the question that which view we should trust for unlabeled data during co-training. Bad predictions from one view may hurt the training procedure of other views through pseudo-label assignments. Meanwhile, encouraging to trust a good prediction as a ``strong'' label from co-training will boost the performance, and lead to improved performance of overall semi-supervised learning. Instead of assigning a pseudo-label for each view directly from the predictions of other views, we propose an adaptive approach, namely \textit{uncertainty-weighted label fusion module} (ULF), to fuse the outputs of different views. ULF is built up of all the views, takes the predictions of each view as input, and then outputs a set of pseudo labels for each view.

Motivated by the uncertainty measurements in Bayesian deep networks, we measure the uncertainty of each view branch for each training sample after turning our model into a Bayesian deep network by adding dropout layers. Between the two types of uncertainty candidates -- aleatoric and epistemic uncertainties, we choose to compute the epistemic uncertainty that is raised by not having enough training data\cite{kendall2017uncertainties}. Such measurement fits the semi-supervised learning goal: to improve model generalizability by exploring unlabeled data. Suppose $y$ is the output of a Bayesian deep network, then the epistemic uncertainty can be estimated by the following equation:

\begin{equation}
    U_e(y) \approx \frac{1}{K} \sum_{k = 1}^K \hat{y_k}^2 - (\frac{1}{K}\sum_{k=1}^K\hat{y_k})^2,
\end{equation}
where $\{ \hat{y}_k\}_{k=1}^K$ are a set of sampled outputs.

With a transformation function $\mathbf{h(\cdot)}$, we can transform the uncertainty score into a confidence score $\mathbf{c}(y) = \mathbf{h}(U_e(y))$. After normalization over all views, the confidence score will act as the weight for each prediction to assign as a pseudo label for other views. The pseudo label $\hat{\mathbf{Y_i}}$ assigned for a single view $i$ can be formatted as 
\begin{equation}
	\hat{\mathbf{Y_i}} = \frac{\sum_{j\neq i}^N\mathbf{c}(p_j(\mathbf{X}))p_j(\mathbf{X})}{\sum_{j\neq i}^N\mathbf{c}(p_j(\mathbf{X}))}.
\label{Eqn:pseudo-label}
\end{equation}

\subsection{Implementation Details}
\paragraph{Network structure.}
In practice, we build an encoder-decoder network based on ResNet-18\cite{he2016deep}, and modified it into a 3D version. For the encoder part, the first $7\times 7$ convolutional layer is inflated into $7\times 7 \times 3$ kernels for low-level 3D feature extraction, similar to \cite{liu20173d}. All other $3\times 3$ convolutional layers are simply changed into $3\times 3\times 1$ that can be trained as a 3D convolutional layer. In the decoder part, we adopt 3 skip connections from the encoder followed by 3D convolutions to give low-level cues for more accurate boundary prediction needed in segmentation tasks.

\vspace{-0.3cm}
\paragraph{Uncertainty-weighted label fusion.}
In terms of view confidence estimation, we modify the network into a Bayesian deep network by adding dropout operations. We sample $K = 10$ outputs for each view and compute voxel-wise epistemic uncertainty. Since the voxel-wise uncertainty can be inaccurate, we sum over the whole volume to finalize the uncertainty for each view. We simply use the reciprocal for the confidence transformation function $\mathbf{h}(\cdot)$ to compute the confidence score. The pseudo label assigned for one view is a weighted average of all predictions of multiple views based on the normalized confidence score. 

\vspace{-0.3cm}
\paragraph{Data pre-processing.}
All the training and testing data are firstly re-sampled to an isotropic volume resolution of 1.0 $mm$ for each axis. Data intensities are normalized to have zero mean and unit variance. We adopt patch-based training, and sample training patches of size $96^3$ with $1$:$1$ ratio between foreground and background. Unlike other 3D segmentation approaches, our approach does not rely on any kind of 3D data augmentation due to the effectiveness of initialization with 2D pre-trained weights.

\vspace{-0.3cm}
\paragraph{Training.}
The used training algorithm is shown in Algorithm~\ref{algo1}. Note that under the semi-supervised setting, the co-training loss is only minimized on the unlabeled data. It is not applied to labeled data as the segmentation loss is already optimized to force the network's prediction to be close to the ground truth.
However, we will later show that the co-training loss can also help each sub-network to learn better features on labeled data. 
The Dice loss \cite{milletari2016v} is used as the segmentation loss function. It performs robustly with imbalanced training data and mitigates the gap between the training objective and commonly used evaluation metrics, such as Dice score.

We firstly train the views separately on the labeled data and then conduct our co-training by fine-tuning the weights. The stochastic gradient descent (SGD) optimizer is used in both stages. In the view-wise training stage, a constant learning rate policy at $7\times 10^{-3}$, momentum at $0.9$ and weight decay of $4\times 10^{-5}$ for 20k iterations is used. In the co-training stage, we adopt a constant learning rate policy at $1\times 10^{-3}$, with the parameter $\lambda_{cot} = 0.2$ and train for 5k iterations. The batch size is 4 in both stages. Our framework is implemented in PyTorch. The whole training procedure takes $\sim$12 hours on 4 NVIDIA Titan V GPUs.

\vspace{-0.3cm}
\paragraph{Testing.}
In the testing phase, there are two choices to finalize the output results: either to choose one single view prediction or to ensemble the predictions of the multi-view outputs. We will report both results in the following sections for a fair comparison with the baselines since the multiple view networks can be thought of being similar to the ensemble of several single view models. The experimental results show that our model improves the performance in both settings (single view and multi-view ensemble). We use sliding-window testing and re-sample our testing results back to the original image resolution to obtain the final results. Testing time for each case ranges from $1$ minute to $5$ minutes depending on the size of the input volume.

\section{Experiments}
In this section, our framework is tested on two popular organ segmentation datasets: NIH pancreas segmentation datasets ~\cite{roth2015deeporgan} and LiTS liver tumor segmentation dataset under semi-supervised settings. Moreover, noticing that our approach is also applicable to fully-supervised settings, we apply it to supervised training and show the benefits even when all the training data is labeled. 
\subsection{Semi-supervised Segmentation}

\subsubsection{NIH Pancreas Segmentation Dataset}
The NIH pancreas segmentation dataset contains 82 abdominal CT volumes. The width and height of each volume are 512, while the axial view slice number can vary from 181 to 466. Under semi-supervised settings, the dataset is randomly split into 20 testing cases and 62 training cases. We report the results of 10\% labeled training cases (6 labeled and 56 unlabeled), 20\% labeled training cases (12 labeled and 50 unlabeled) and 100\% labeled training cases. In the results, the performance of one single view (the average of all single views' DSC scores) is reported for a fair comparison, not a multi-view ensemble (see Table ~\ref{Tab:semiTE}).

\begin{table}[!btp]
\centering
\begin{tabular}{|l|c|c|c|c|}    
\hline
Method & Backbone  & 10\% lab& 20\% lab\\
\hline\hline
Supervised & 3D ResNet-18  & 66.75 & 75.79\\
\hline\hline
DMPCT~\cite{zhou2018semi}  &  2D ResNet-101 & 63.45& 66.75\\
DCT~\cite{qiao2018deep} (2v) & 3D ResNet-18 & 71.43& 77.54\\
TCSE~\cite{li2018semi} &  3D ResNet-18 & 73.87 & 76.46\\
\hline\hline
Ours (2 views) &  3D ResNet-18 &\textbf{75.63} & \textbf{79.77}\\
Ours (3 views) &  3D ResNet-18 &\textbf{77.55} & \textbf{80.14}\\
Ours (6 views) &  3D ResNet-18 &\textbf{77.87} & \textbf{80.35}\\
\hline
\end{tabular}
\vspace{-0.2cm}
\caption{
    Comparison to other semi-supervised approaches on NIH dataset (DSC, \%).  Note that we use the same backbone network as~\cite{li2018semi}~\cite{qiao2018deep}. Here, ``2v" means two views. For our approach, the average of all single views' DSC score is reported for a fair comparison, not a multi-view ensemble. ``10\% lab"  and ``20\% lab" mean the percentage of labeled data used for training.
}
\label{Tab:semiTE}
\vspace{-0.6cm}
\end{table}

\begin{figure}
\begin{center}
    \includegraphics[width=8.5cm]{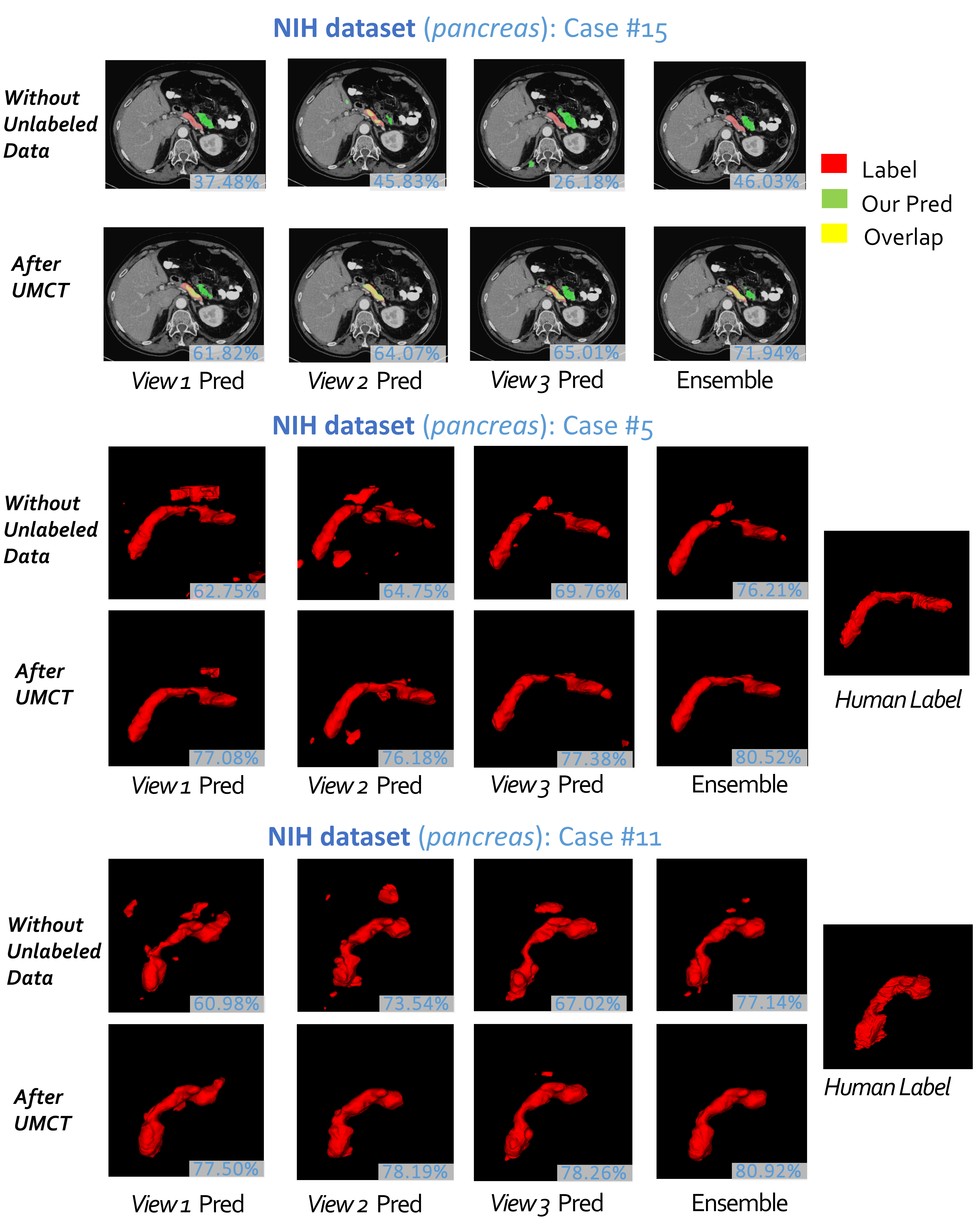}
\end{center}
\vspace{-0.5cm}
\caption{
    2D and 3D visualizations for 3 cases in the test set under 10\% labeled data setting. DSC score is largely improved by our co-training approach. Best viewed in color.
}
\label{Fig:semi}
\vspace{-0.5cm}
\end{figure}

The segmentation accuracy is evaluated by Dice-S{\o}rensen coefficient (DSC). A large margin improvement over the fully supervised baselines in terms of single view performance can be observed, proving that our approach effectively leverages the unlabeled data. A Wilcoxon signed-rank test comparing to the supervised baseline's results (20\% labeling) shows significant improvements of our approach with a $p$-value of 0.0022. Fig.~\ref{Fig:semi} shows 3 cases in 2D and 3D with ITK-SNAP ~\cite{py06nimg}. In addition, our model is compared with the state-of-the-art semi-supervised approach of deep co-training ~\cite{qiao2018deep} and recent semi-supervised medical segmentation approaches. In particular, we compare to Li et al. ~\cite{li2018semi} who extended the $\pi$ model ~\cite{laine2016temporal} with transformation consistent constraints; and Zhou et al. ~\cite{zhou2018semi} who extended the self-training procedure by iteratively updating pseudo labels on unlabeled data using a fusion of three 2D networks trained on cross-sectional views. The results reported in Tab.~\ref{Tab:semiTE} are based on our careful re-implementations in order to allow a fair comparison.

Our implementations of ~\cite{qiao2018deep} and ~\cite{li2018semi} are operated on the axial view of our single view branch with the same backbone structure (our customized 3D ResNet-18 model). Our co-training approach achieve about 4\% gain in the 10\% labeled and 90\% unlabeled settings. We also find that improvements of other approaches are small in the 20\% settings (only 1\% compared to the baseline), while ours still is capable to achieve a reasonable performance gain with the growing number of labeled data. For ~\cite{zhou2018semi} with a 2D approach, their experiment is conducted on 50 labeled cases. We modify their backbone network (FCN ~\cite{long2015fully}) into DeepLab v2 ~\cite{chen2018deeplab}, in order to fit our stricter settings (6 and 12 labeled cases). This modification leads to an improvement of 3\% in 100\% fully supervised training (from 73\% to 76\%). Their approach outputs the result after using an ensemble of three models. 

Since the main difference in two-view learning between our approach and ~\cite{qiao2018deep} is the way of encouraging view differences, the results illustrate the effectiveness of our multi-view analysis combined with asymmetric feature learning on 3D co-training. With more views, our uncertainty-weighted label fusion can further improve co-training performance. We will report ablation studies on it in section~\ref{sec:ablation}.

\begin{figure}[H]
\begin{center}
    \includegraphics[height=4.6cm]{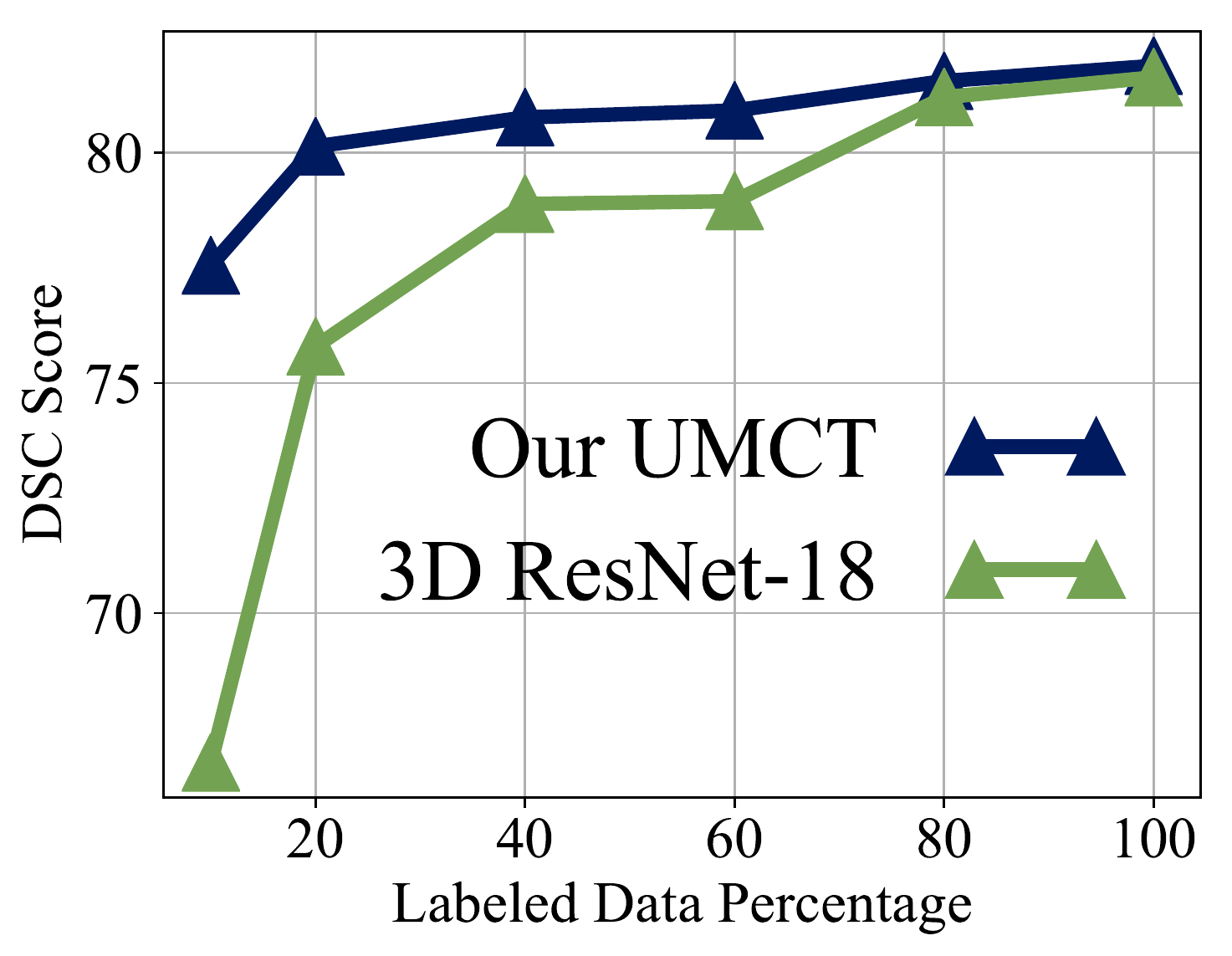}
\end{center}
\vspace{-0.6cm}
\caption{
    Performance plot of our semi-supervised approach over the fully-supervised baseline on different labeled data ratio.
}
\label{Fig:percentage}
\end{figure}

Furthermore, we performed a study on data utilization efficiency of our approach compared to the baseline fully-supervised network (3D ResNet-18). Fig. \ref{Fig:percentage} shows the performance change according to labeled data proportion on NIH pancreas segmentation. From the plot, it can be seen that when labeled data is over 80\%, simple supervised training (with 3D ResNet-18) suffices. Note that our approach with 20\% labeled data (DSC 80.35\%) performs better than 60\% supervised training (DSC 78.95\%). At such a percentage, our approach can save $\sim$ 70\% of the labeling efforts.

\vspace{-0.1cm}

\subsubsection{LiTS Liver Tumor Segmentation Challenge}

We also report our results on the training set of LiTS Liver Tumor Segmentation Challenge. The 131 cases are randomly split into 100 training and 31 testing cases. The input volumes are all abdominal CT scans. The segmentation target contains 2 classes: liver (large and less challenging) and lesion (tumors with large variance in size, more challenging). Our semi-supervised settings are the same as those used in NIH pancreas dataset experiments. We report results on 10\% labeled data (10 labeled cases and 90 unlabeled cases) and 20\% labeled data (20 labeled case and 80 unlabeled cases) with 3-view co-training. The performance of single view and multi-view ensemble both improves as shown in Table~\ref{Tab:semiLiTS}. The improvement on liver segmentation is limited (less than 1\%) because the liver segmentation is already very good with a single view only. If we only use 10\% data for supervised training, we can already reach 93.17\% after fusing the three views' results by majority voting. However, we see a large margin improvement in the more challenging lesion segmentation, especially under ``our UMCT 20\%" settings (even more than ``our UMCT 10\%"). We hypothesize that the case variance of lesions is larger than normal organs (pancreas, liver, etc).  With only 10 cases for our labeled set, $\mathcal{L}_{sup}$ can misguide the training procedure and introduce bias to the labeled set (known as overfitting). However, using $\mathcal{L}_{cot}$ to explore the unlabeled part of the dataset, we can train a more robust model compared to fully supervised training using the same number of labeled cases. Overall, the improvements are significant even on the challenging liver lesions with large case variance.

\begin{table}[!btp]
\centering
\begin{tabular}{l|c|c|c|c|}
\cline{2-5}
& \multicolumn{2}{c}{Liver}& \multicolumn{2}{|c|}{Lesion}\\
\hline
\multicolumn{1}{|l|}{Method}  &  Single & \multicolumn{1}{c|}{MV}& Single & \multicolumn{1}{c|}{MV}\\
\hline\hline
\multicolumn{1}{|l|}{100\% Supervised} & 95.07&	\multicolumn{1}{c|}{95.50}& 64.00&	\multicolumn{1}{c|}{65.65}\\
\hline\hline
\multicolumn{1}{|l|}{10\% Supervised} & 92.23 &	\multicolumn{1}{c|}{93.17}& 43.98&	\multicolumn{1}{c|}{48.90}\\
\multicolumn{1}{|l|}{20\% Supervised} & 93.06 &	\multicolumn{1}{c|}{94.52}& 50.39&	\multicolumn{1}{c|}{53.15}\\
\hline\hline
\multicolumn{1}{|l|}{our UMCT 10\%} &  92.98&	\multicolumn{1}{c|}{93.53}& 49.79&	\multicolumn{1}{c|}{52.14}\\
\multicolumn{1}{|l|}{our UMCT 20\%} &94.40&	\multicolumn{1}{c|}{94.81}& 57.76&	\multicolumn{1}{c|}{59.60}\\
\hline
\end{tabular}
\vspace{-0.2cm}
\caption{
    Our 3-view co-training on LiTS dataset (DSC, \%). ``Single" means the DSC score of one single view, while ``MV" means multi-view ensemble. The first three rows are our fully supervised baselines. The last two rows are the results of our approach, with 10\% labeled data and 20\% labeled data. We report both liver and lesion (tumor) results. The improvements using UMCT over the corresponding baselines are significant, especially for the performance on liver lesions.
}
\label{Tab:semiLiTS}
\end{table}

\begin{table}[!btp]
\centering
\begin{tabular}{|c|c|c|}    
\hline
Method & Liver & Lesion \\
\hline\hline
3D AH-Net ~\cite{liu20173d} & \textbf{96.3} & 63.4\\
H-DenseUNet~\cite{li2017h} & 96.1 & 72.2 \\
\hline\hline
3 views UMCT (ours) & 95.9 & \textbf{72.6} \\
\hline
\end{tabular}
\vspace{-0.2cm}
\caption{
    Results of fully supervised training with UMCT on LiTS test set (DSC, \%).
}
\vspace{-0.5cm}
\label{Tab:LiTSFull}
\end{table}



\subsection{Application to Fully Supervised Settings}
Our approach can also be applied to fully supervised training. On semi-supervised tasks, we do not see a clear improvement when enforcing $\mathcal{L}_{cot}$ on labeled data because of the quantity limitation. However, when labeled data is sufficient, we want to see if our multi-view co-training can guide each 2D-initialized branch to help each other by enforcing 3D consistency. The final framework for fully supervised training is: we firstly train the sub-networks of different views separately, and then fine-tune with the following loss function:
\vspace{-0.2cm}
\begin{equation}
\mathcal{L} = \sum_{(\mathbf{X},\mathbf{Y}) \in \mathcal{S}} [\mathcal{L}_{sup}(\mathbf{X}, \mathbf{Y}) + \lambda \mathcal{L}_{cot}(\mathbf{X})]
\label{Eqn:Full}
\vspace{-0.2cm}
\end{equation}

\begin{table}[!btp]
\centering
\begin{tabular}{|p{2.5cm}|cc|cc|}    
\hline
Task & \multicolumn{2}{c|}{DSC} & \multicolumn{2}{c|}{NSD}\\
\hline\hline
Hepatic Vessel & 0.63 & 0.64 & 0.83 & 0.72\\
Spleen & \multicolumn{2}{l|}{0.96}& \multicolumn{2}{l|}{1.00}\\
Colon & \multicolumn{2}{l|}{0.56} & \multicolumn{2}{l|}{0.66} \\
\hline
\end{tabular}
\vspace{-0.2cm}
\caption{
    DSC and NSD (normalized surface distance) scores on the final validation phase of Medical Segmentation Challenge (some tasks have multi-class labels).
}
\vspace{-0.2cm}
\label{Tab:Decathlon}
\end{table}

On LiTS dataset challenge, a fully-supervised method based on our 3-view co-training method achieved the state-of-the-art results in terms of tumor segmentation DSC score and comparable liver segmentation results, see Table~\ref{Tab:LiTSFull}.

On Medical Segmentation Decathlon challenge, a fully-supervised method based on our 3-view co-training method achieved the second place in the final testing phase, see Tabel~\ref{Tab:Decathlon}. One goal of the challenge was that without any hyperparameter change allowed, a favored model has to be generalizable and robust to various segmentation tasks. Our model can satisfy such requirements because we have the following features. First, our model, although trained on 3D patches, is initialized from 2D pre-trained models. We will further discuss the influence of 2D pre-trained models in the next section. Second, we have three views of networks and use $\mathcal{L}_{cot}$ to help each other gaining more 3D information through the multi-view co-training process. These two characteristics boost the robustness of our model on supervised volumetric segmentation tasks.

\begin{table*}
\begin{subtable}[t]{0.48\textwidth}
\begin{tabular}{|l|c|r|r|}    
\hline
Backbone & Params &Sup 10\% & Semi 10\%\\
\hline\hline
VNet & 9.44M& 66.97 & 76.89	\\
3D ResNet-18 &11.79M &66.76& \textbf{77.55}\\
3D ResNet-50 & 27.09M&67.96& \textbf{78.74}\\
\hline
\end{tabular}
\caption{\footnotesize Ablation studies on backbone structures (3 views UMCT).}
\label{tab:ab_backbone}
\end{subtable}
\begin{subtable}[t]{0.48\textwidth}
\vspace{-1cm}
\begin{tabular}{|l|r|r|r|r|}    
\hline
Method & Coronal & Sagittal & Axial & MV\\
\hline\hline
100\% Supervised & 82.13 & 81.41 & 82.53 & 84.18\\
\hline
UMCT Supervised &  \textbf{82.61} & \textbf{82.35} & \textbf{83.44} & \textbf{84.61}\\
\hline
\end{tabular}
\caption{
    \footnotesize Our UMCT on 100\% labeled data from the NIH data. The first row is pure single view training, while the second is UMCT. ``Coronal", ``Sagittal" and ``Axial" correspond to three views in CT scan in radiology.
}
\label{tab:ab_sup}
\end{subtable}
\hspace{\fill}
\begin{subtable}[t]{0.33\linewidth}
\centering
\begin{tabular}{|l|c|}    
\hline
Views & DSC(\%)\\
\hline\hline
2 views & 75.63\\
\hline
3 views & 76.49\\
3 views + ULF & \textbf{77.55}\\
\hline
6 views & 76.94\\
6 views + ULF& \textbf{77.87}\\
\hline
\end{tabular}
\caption{
    \footnotesize On uncertainty-weighted label fusion (ULF) with difference views in training (10\% labeled data, 3D ResNet-18).
}
\label{tab:ab_ULF}
\end{subtable}
\hspace{0.3cm}
\begin{subtable}[t]{0.25\linewidth}
\centering
\begin{tabular}{|l|c|}    
\hline
$\lambda_{cot}$ & DSC(\%)\\
\hline\hline
0.1 &   77.28\\
\textbf{0.2} &   \textbf{77.55}\\
0.5 &   77.38\\
\hline
\end{tabular}
\vspace{0.4cm}
\caption{
    \footnotesize $\lambda_{cot}$(10\% labeled data, 3 views, 3D ResNet-18).
}
\label{tab:ab_lambda}
\end{subtable}
\hspace{\fill}
\begin{subtable}[t]{0.37\textwidth}
\centering
\begin{tabular}{|l|r|r|}    
\hline
Model & w/o Init & w/ Init\\
\hline\hline
Deeplab-3D &76.09& 80.11\\
Our 3D ResNet-50 &78.70& 82.53\\
\hline
\end{tabular}
\vspace{0.58cm}
\caption{
    \footnotesize On the influence of initialization for 3D models. Experiments are done on axial view, NIH dataset.
}
\label{tab:ab_init}
\end{subtable}
\vspace{-0.5cm}
\caption{Ablation studies for our UMCT on NIH dataset.}
\label{tab:ablation}
\end{table*}

\subsection{Ablation Studies}
\label{sec:ablation}
In this subsection, we will provide several ablation studies for each component of the proposed UMCT framework.

\vspace{-0.4cm}
\paragraph{On the backbone network structure.}
Our backbone selection (2D-initialized, heavily asymmetric 3D architecture) will introduce 2D biases in the training phase while benefiting from such 2D pre-trained models. We have claimed that we can utilize the complementary information from 3-view networks while exploring the unlabeled data with UMCT. We give an ablation study on the network structure, which contains a V-Net \cite{milletari2016v}, a common 3D segmentation network with all symmetrical kernels in all dimensions. Such network also shares a similar amount of parameters with our customized 3D ResNet-18, see Table~\ref{tab:ab_backbone}. The results of V-Net show that our multi-view co-training can be generally and successfully applied to 3D networks. Although the results of fully supervised parts are similar, our ResNet-18 outperforms V-Net by more than 1\%, illustrating that our asymmetric design, encouraging view differences, brings advantages over traditional 3D deep networks.

\vspace{-0.4cm}
\paragraph{On uncertainty-weighted label fusion (ULF), number of views and parameter $\lambda_{cot}$.}
ULF acts as an important role in pruning out bad predictions and keeping good ones as supervision to train other views. Table~\ref{tab:ab_ULF} gives 
the single view results in multiple views experiments.  
The performance becomes better with more views. For two views, ULF is not applicable since we can only obtain one view prediction as a pseudo label for the other view. For three views and six views, ULF helps boost the performance, illustrating the effectiveness of our proposed approach for view confidence estimation. We also tried different values of $\lambda_{cot}$ in Table~\ref{tab:ab_lambda}, where performance variance is not large. We choose $\lambda_{cot}=0.2$ in our experiments.

\vspace{-0.4cm}
\paragraph{On fully supervised training.}
Table~\ref{tab:ab_sup} shows how our multi-view co-training helps with the fully supervised training on the NIH dataset. The model used is our 3D ResNet-50 with 3 views co-training. Our approach improves the results on each single model, as well as the multi-view ensemble results.

\vspace{-0.4cm}
\paragraph{On network initialization.}
We address the importance of initialization for training a robust 3D model. This subsection provides an ablation study on the influence of initialization of 3D networks in the field of 3D segmentation, which is often neglected by previous works. We trained two 3D ResNet-50 (in our settings) in axial view on the NIH dataset with all 100\% labeled data. Here, one model uses 2D initialization, while the other is trained from scratch. We also conduct similar comparisons with a DeepLab-3D model, where we directly change each 2D kernel of DeepLab(v2)-ResNet101 model into a 3D kernel. We initialize DeepLab-3D in the same way as ~\cite{carreira2017quo}. Table~\ref{tab:ab_init} shows the comparison. Those models with initialization perform remarkably better. Thus, we believe that initialization is helpful to train 3D models for volumetric segmentation. Using weights from the pre-trained models of natural image tasks is beneficial for learning process. It would be a promising research direction to investigate approaches on 3D network initialization or providing 3D models pre-trained on large-scale datasets.

\vspace{-0.2cm}
\section{Conclusion}
\vspace{-0.2cm}
In this paper, we presented \textit{uncertainty-aware multi-view co-training} (UMCT), aimed at 3D semi-supervised learning. We extended dual view co-training and deep co-training on 2D images into multi-view 3D training, naturally introducing data-level view differences. We also proposed asymmetrical 3D kernels initialized from 2D pre-trained models to introduce feature-level view differences. In multi-view settings, an \textit{uncertainty-weighted label fusion module} (ULF) is built to estimate the accuracy of each view prediction by Bayesian uncertainty measurement. Epistemic uncertainty was estimated after transforming our model into a Bayesian deep network by adding dropout. This module gives a larger weight to more confident predictions and further boost the performance on multi-view predictions. Experiments under semi-supervised setting were performed on the NIH pancreas dataset and the LiTS liver tumor dataset. Other approaches were outperformed by a large margin on the NIH dataset. We also applied co-training objectives on labeled data under fully supervised settings. The results were also promising, illustrating the effectiveness of multi-view co-training on 2D-initialized networks.

\vspace{-0.3cm}
\paragraph{Acknowledgements} We thank Dr. Lingxi Xie, Siyuan Qiao and Yuyin Zhou for instructive discussions.

{\small
\bibliographystyle{ieee}
\bibliography{egbib}
}

\end{document}